\documentclass[10pt,twocolumn,letterpaper]{article}

\usepackage{iccv}
\usepackage{times}
\usepackage{graphicx}
\graphicspath{{figures/}}
\usepackage{subfig}
\usepackage{amsmath}
\usepackage{amssymb}
\usepackage{multirow}
\usepackage{booktabs}
\usepackage[table]{xcolor}
\def\pt{\phantom{0}}
\usepackage{bbding}
\definecolor{LightCyan}{rgb}{0.83,0.95,1}
\definecolor{grassgreen}{rgb}{0.1,0.8,0.2}
\usepackage{threeparttable}
\newcommand{\squishlist}{
 \begin{list}{$\bullet$}
  { \setlength{\itemsep}{0pt}
     \setlength{\parsep}{1pt}
     \setlength{\topsep}{1pt}
     \setlength{\partopsep}{0pt}
     \setlength{\leftmargin}{0.8em}
     \setlength{\labelwidth}{1em}
     \setlength{\labelsep}{0.5em} } 
}

\newcommand{\squishend}{
  \end{list}  
}

\usepackage[pagebackref=true,breaklinks=true,colorlinks,bookmarks=false]{hyperref}

\iccvfinalcopy 

\ificcvfinal\fi

\begin{document}

\title{On Improving the Algorithm-, Model-, and Data- Efficiency of Self-Supervised Learning}

\author{
Yun-Hao Cao, Jianxin Wu\thanks{J. Wu is the corresponding author.}
\\
National Key Laboratory for Novel Software Technology, Nanjing University, China\\
\tt caoyh@lamda.nju.edu.cn, \tt wujx2001@nju.edu.cn
}

\maketitle
\ificcvfinal\thispagestyle{empty}\fi

\begin{abstract}
   Self-supervised learning (SSL) has developed rapidly in recent years. However, most of the mainstream methods are computationally expensive and rely on two (or more) augmentations for each image to construct positive pairs. Moreover, they mainly focus on large models and large-scale datasets, which lack flexibility and feasibility in many practical applications. In this paper, we propose an efficient single-branch SSL method based on non-parametric instance discrimination, aiming to improve the algorithm, model, and data efficiency of SSL. By analyzing the gradient formula, we correct the update rule of the memory bank with improved performance. We further propose a novel self-distillation loss that minimizes the KL divergence between the probability distribution and its square root version. We show that this alleviates the infrequent updating problem in instance discrimination and greatly accelerates convergence. We systematically compare the training overhead and performance of different methods in different scales of data, and under different backbones. Experimental results show that our method outperforms various baselines with significantly less overhead, and is especially effective for limited amounts of data and small models. 
\end{abstract}

\section{Introduction}

Deep supervised learning has achieved great success in the last decade. However, traditional supervised learning approaches rely heavily on a large set of annotated training data. Self-supervised learning (SSL) has gained popularity because of its ability to avoid the cost of annotating large-scale datasets as well
as the ability to obtain task-agnostic representations. After the emergence of
the contrastive learning (CL) paradigm~\cite{InfoNCE:arxiv2018, simclr:hinton:ICML20}, SSL has clearly gained momentum and several recent works~\cite{mocov2:xinlei:arxiv2020,byol:grill:NIPS20,swav:caron:NIPS20} have achieved comparable or even better accuracy than the supervised pertaining when transferring to downstream tasks. However, these methods are almost all dual-branched, that is, the network needs to generate at least two views for each image during learning. What's worse, the combination of a time-consuming algorithm (dual-branched), a large-scale dataset (e.g., ImageNet), a complex backbone (e.g., ResNet-50), and a large number of epochs (800 or more) means that SSL methods are computationally extremely expensive. This phenomenon makes SSL a privilege for researchers at few institutions. In this paper, we propose to improve the efficiency of SSL methods from three aspects: algorithm (training) efficiency, model efficiency, and data efficiency.

\begin{figure}
    \centering
    \includegraphics[width=0.9\linewidth]{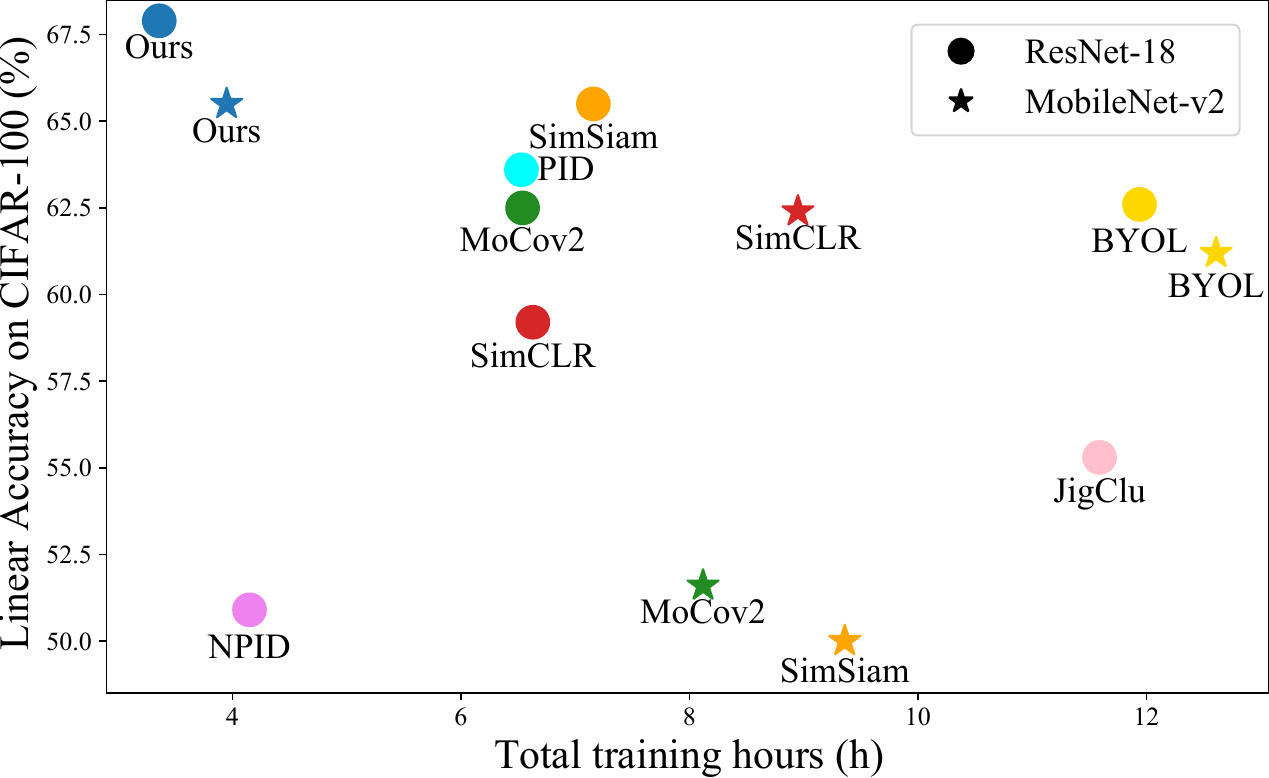}
    \caption{Linear probing accuracy and training cost (in hours) of different SSL methods on CIFAR-100~\cite{cifar}.}
    \label{fig:performance}
\end{figure}

As an alternative to dual-branch SSL, single-branch methods~\cite{deepclustering:caron:ECCV18,rotnet:spyros:ICLR18,jiasaw:mehdi:ECCV16} only require a single crop for each image in each iteration, which naturally reduces the training overhead per iteration. As a representative of them, parametric instance discrimination methods~\cite{exemplar:alexey:nips14,onemillion:liu:AAAI21,idmm:ECCV2022} learn to classify every example into its own category. However, the final parametrized classification layer will bring an intolerable increase in computation and GPU memory usage as the number of training data increases. As a solution, NPID~\cite{memorybank:wu:CVPR18} transforms instance discrimination into a non-parametric version by maintaining a memory bank but its accuracy is far behind mainstream contrastive learning methods. MoCo~\cite{moco:kaiming:CVPR20} improves NPID using a momentum encoder at the cost of turning to dual-branch again. Chen \etal ~\cite{jigclu:CVPR21} proposed a jigsaw clustering task to improve single-branch SSL but the complicated pipeline makes its training overhead even larger than many dual-branch methods. Therefore, how to design an efficient and effective single-branch self-supervised method is challenging.

In this paper, we aim to bridge the accuracy gap between single- and dual-branch methods while maintaining the training efficiency of single-branch methods. Our method is based on NPID~\cite{memorybank:wu:CVPR18}, but with the following three important improvements. First, we perform a forward pass on the untrained network to obtain features as the initialization of the memory bank, which was randomly initialized in both NPID and MoCo. Inspired by \cite{byol:grill:NIPS20,onemillion:liu:AAAI21}, we know that a randomly initialized network also has representation ability, and experiments show that our initialization can speed up the convergence with negligible cost. Second, we revise the update rule of the memory bank based on gradient formulation. In \cite{memorybank:wu:CVPR18}, the feature of the $i$-th instance will only be used to update the weights of the $i$-th class. By analyzing the weights' gradient, we know that the feature of an instance will also be passed back to update the weights corresponding to other instances using our update rule. Third, we design an effective self-distillation loss that minimizes the KL divergence of the probability distribution and the distribution after taking the square root. Theoretical and empirical results demonstrate that this loss can effectively solve the problem of infrequent updating~\cite{idmm:ECCV2022} in instance discrimination and greatly accelerate convergence, achieving better performance with less overhead, as shown in Fig.~\ref{fig:performance}.

In addition to improving algorithm efficiency, we also try to improve the model and data efficiency in self-supervised learning. In practical applications, many models need to be deployed on terminal devices with limited memory, computation, and storage capabilities. Hence, self-supervised learning with small models is an important problem. Fang \etal~\cite{seed:fang:ICLR21} found that small models perform poorly under the paradigm of self-supervised contrastive learning and smaller models with fewer parameters cannot effectively learn instance-level discriminative representation with a large amount of data. SEED~\cite{seed:fang:ICLR21} and DisCo~\cite{disco:ECCV2022} adopt knowledge distillation to address this problem and Shi \etal~\cite{smallmodel:AAAI22} tweaked hyperparameters and image augmentations to improve performance on small models. In this paper, we show that our method can effectively improve the performance of small models and speed up the convergence of instance discrimination tasks for them. 

From the perspective of data efficiency, many realistic scenarios require that we cannot always rely on large-scale training data. For example, it is difficult to collect large-scale training data in some fields (e.g., medical images). Also, fast model iteration (e.g., update a model in 10 minutes) forbids us from using large-scale data for training. Therefore, in this paper, we study the performance of different SSL methods under different scales of training data. Experimental results demonstrate the data efficiency of our method, and our improvements will increase as the amount of data decreases. In summary, our contributions are:

\squishlist
    \item We propose a single-branch method, which improves the training efficiency, model efficiency, and data efficiency of self-supervised learning.
    \item We propose the initialization method of the memory bank, and revise the update rule based on the gradient formula.
    \item We propose a self-distillation KL loss to alleviate the infrequent updating problem for instance discrimination, which greatly accelerates the convergence. 
    \item We systematically compare the efficiency of different SSL methods, and exhaustive experiments show that our method achieves better performance on various benchmarks with less training overhead. Moreover, our method is extremely effective for lightweight models and small data, and our advantages will be further amplified as the amount of data decreases.
\squishend

\begin{figure*}
    \centering
    \includegraphics[width=0.8\linewidth]{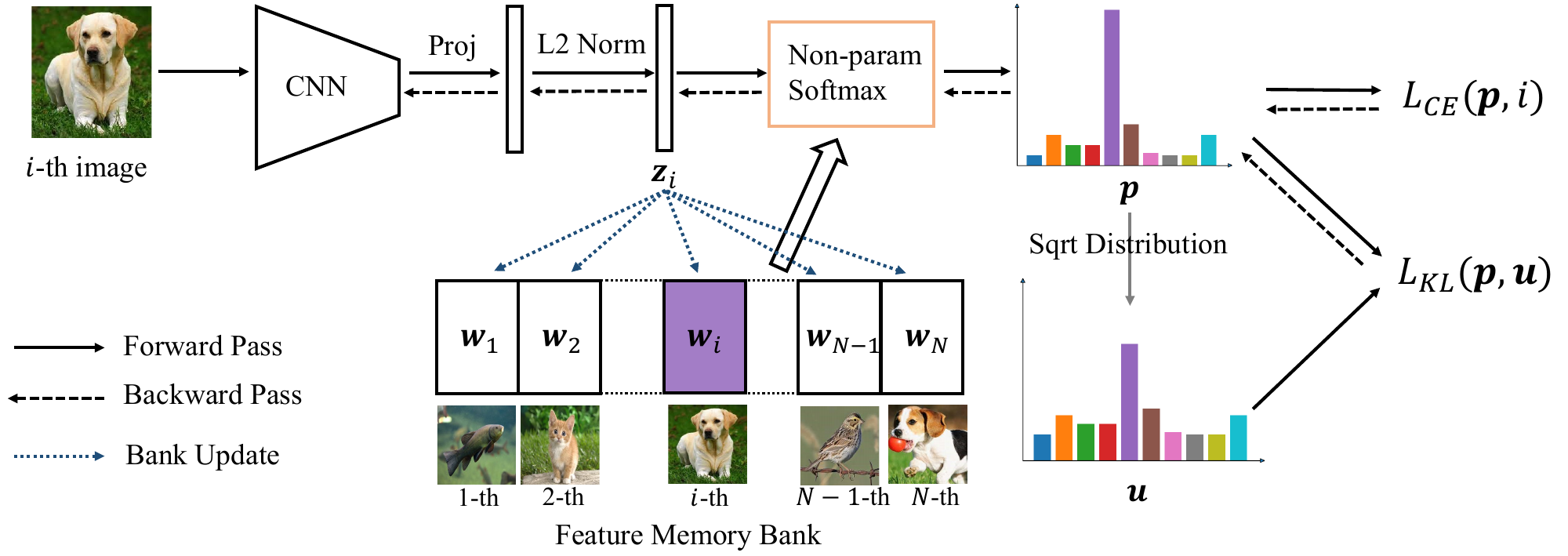}
    \caption{The general framework of our method.}
    \label{fig:network}
\end{figure*}

\section{Related Works}

Self-supervised learning (SSL) has emerged as a powerful method to learn visual representations without labels.
Many recent works follow the contrastive learning paradigm~\cite{InfoNCE:arxiv2018}. For instance, SimCLR~\cite{simclr:hinton:ICML20} and MoCo~\cite{moco:kaiming:CVPR20} train networks to identify a pair of views originating from the same image when contrasted with many views from other images. Follow-up works BYOL~\cite{byol:grill:NIPS20} and SimSiam~\cite{simsiam:kaiming:cvpr2021} discard negative sampling in contrastive learning but achieve even
better results using siamese networks. Unlike the siamese structure in contrastive methods, single-branch methods~\cite{rotnet:spyros:ICLR18, jiasaw:mehdi:ECCV16, deepclustering:caron:ECCV18, exemplar:alexey:nips14} propose different pretext tasks to train unsupervised models. Pretext-based approaches mainly explore the context features of images or videos such as context similarity~\cite{jiasaw:mehdi:ECCV16, context:carl:ICCV15}, spatial structure~\cite{rotnet:spyros:ICLR18}, clustering property~\cite{deepclustering:caron:ECCV18}, temporal structure~\cite{sorting:lee:ICCV17}, etc. Parametric instance discrimination~\cite{exemplar:alexey:nips14,ParametricInstance:cao:arxiv2020,onemillion:liu:AAAI21} learns to discriminate between a set of surrogate classes, where each class represents different transformed patches of a single image. NPID~\cite{memorybank:wu:CVPR18} employs non-parametric instance discrimination by maintaining a memory bank but its performance is far behind the mainstream contrastive learning methods. JigClu~\cite{jigclu:CVPR21} improves the performance of single-branch methods at the cost of greater training overhead.

There are also some recent works trying to improve the efficiency of SSL in different dimensions. SEED~\cite{seed:fang:ICLR21} and DisCo~\cite{disco:ECCV2022} study self-supervised learning with small models. SSQL~\cite{ssql:cao:ECCV2022} proposes to pretrain quantization-friendly self-supervised models to facilitate downstream deployment. Cao \etal~\cite{S3L:cao:arxiv2021} and Cole \etal~\cite{does:CVPR22} investigated the data efficiency of self-supervised methods. Fast-MoCo~\cite{fastmoco:ECCV2022} tries to accelerate the training of MoCov2~\cite{mocov2:xinlei:arxiv2020}, which is still a dual-branch method. These previous methods try to improve the SSL efficiency from a single dimension, but we study the efficiency of SSL from three dimensions for the first time in this work.

\section{The Proposed Method}

We begin with the basic notation and a brief introduction of our framework, followed by our algorithm and analysis.

\subsection{Preliminaries}\label{sec:preliminary}

An input image $\boldsymbol{x}_i$ ($i=1,\cdots,N$) is sent to a network $f(\cdot)$ and get output representation $\boldsymbol{z}_i=f(\boldsymbol{x}_i)\in{\mathbb{R}^d}$, where $N$ denotes the total number of instances. Then, a fully connected (FC) layer $\boldsymbol{w}$ is used for classification and the number of classes equals the total number of training images $N$ for parametric instance discrimination. Let us denote the FC's weights as $\boldsymbol{w}_i$ ($i=1,2,\dots,N$), then the prediction for the $i$-th instance is 
\begin{equation}
p_i = \frac{\exp(\boldsymbol{w}_i^T \boldsymbol{z}_i)}{\sum_{j=1}^N \exp(\boldsymbol{w}_j^T \boldsymbol{z}_i)}\,.
\end{equation}
The loss function for the $i$-th instance is 
\begin{equation}
    L_{\text{CE}}=-\log(p_i)\,,
\end{equation}
because every instance is a class and the label for $\boldsymbol{x}_i$ is $i$.

As shown in Fig.~\ref{fig:network}, we use a non-parametric variant following \cite{memorybank:wu:CVPR18}, where each $\boldsymbol{w}_i$ is stored in a feature memory bank without using gradient back-propagation. This eliminates
the need for computing and storing the gradients for $\boldsymbol{w}_i$, improving the storage and training efficiency. 

\subsection{Feature Bank}\label{sec:bank}

Now we describe how we initialize and update the feature memory bank.

\noindent{\textbf{Feature Calibrate.}} NPID~\cite{memorybank:wu:CVPR18} and MoCo~\cite{moco:kaiming:CVPR20} randomly initialize the memory bank while we perform a forward pass on the untrained network to obtain features for initialization, i.e, $\boldsymbol{w}_i=\boldsymbol{z}^{(0)}_i$. This brings negligible overhead, but as we show later in Sec.~\ref{sec:ablation}, speeds up convergence and improves performance.

\noindent{\textbf{Grad Update.}} A naive way to update the weights in the feature bank is to use the current output feature~\cite{memorybank:wu:CVPR18}:
\begin{equation}
    \boldsymbol{w}_i \leftarrow m \boldsymbol{w}_i + (1-m) \boldsymbol{z}^{(t)}_i \,,
\end{equation}
where $\boldsymbol{z}^{(t)}_i$ is the output representation for the $i$-th instance at the $t$-th iteration and $m$ is a hyper-parameter.

However, if we calculate the gradient w.r.t $\boldsymbol{w}_k$:
\begin{equation}
\label{eq:grad}
    \frac{\partial{L_{\text{CE}}}}{\partial{\boldsymbol{w}_k}}=-\delta_{\{k=i\}}\boldsymbol{z}_i+\frac{e^{\boldsymbol{w}_k^T\boldsymbol{z}_i}}{\sum_{j=1}^Ne^{\boldsymbol{w}_j^T{\boldsymbol{z}_i}}}\boldsymbol{z}_i=(p_k-\delta_{\{k=i\}})\boldsymbol{z}_i\,,
\end{equation}
where $\delta$ is an indicator function, equals 1 iff $k=i$.

According to \eqref{eq:grad}, when we sum the loss from all instances, the update direction (i.e., negative gradient) for $\boldsymbol{w}_i$ will be affected by the output of other instances. Specifically, the corrected update direction will be:
\begin{equation}
    \hat{\boldsymbol{z}}_i^{(t)} = (1-p_i)\boldsymbol{z}^{(t)}_i - \sum_{j\neq i} p_j \boldsymbol{z}^{(t)}_j \,.
\end{equation}
Then we use this corrected direction to update the bank:
\begin{equation}
\label{eq:momentum-update}
    \boldsymbol{w}_i \leftarrow m \boldsymbol{w}_i + (1-m) \hat{\boldsymbol{z}}^{(t)}_i \,.
\end{equation}

\subsection{SqrtKL}\label{sec:sqrtkl}

When we do instance discrimination, one important issue is that the updates to FCs are very rare: the gradient with respect to $\boldsymbol{w}_j$ ($j \neq i$) has to be calculated from $-\frac{1}{p_i}$, which is \emph{mostly} related to $p_i$. Now if we back propagate from $p_i$ to $\boldsymbol{w}$, it is \emph{mostly} focused only on updating $\boldsymbol{w}_i$, but \emph{not} other FC weights $\boldsymbol{w}_j$ ($j \neq i$). Although the $\sum_{j=1}^C \exp(\boldsymbol{w}_j^T \boldsymbol{z}_i)$ term involves $\boldsymbol{w}_j$ for $j \neq i$, its impact is negligible in most cases. To be more precise, from \eqref{eq:grad} we know that when $j \neq i$, then the gradient with respect to $\boldsymbol{w}_j$ is $p_j\boldsymbol{z}_i$ --- clearly negligible when $p_j \approx 0$. Or, $\boldsymbol{w}_i$ is updated roughly only once per epoch, thus we need many epochs to converge.

Now we define a square root probability distribution 
$$ u_i = \frac{\sqrt{p_i}}{\sum_{j=1}^N \sqrt{p_j}}$$ for $i=1,2,\dots,N$. $\mathbf{u}=\{u_1,\dots,u_N\}$ will be clearly more balanced than $\mathbf{p}=\{p_1,\dots,p_N\}$, as shown in Fig.~\ref{fig:network}. In addition to the cross entropy loss, we can add a KL divergence loss: 
\begin{equation}
L_{\text{SqrtKL}} = \mathrm{KL}(\mathbf{p}, \mathbf{u})\,.
\end{equation}

Because $\mathbf{u}$ is generated out of $\mathbf{p}$, one network is enough and it is a self-distillation.  Note that ``more balanced'' means even though the prediction $\mathbf{p}$ is very sharp (hence $p_j \approx 0$ if $j \neq i$), $\mathbf{u}$ will be less sharp. One example: let $N=10$, $\mathbf{p}=\{0.91,0.01,\dots,0.01\}$. Then $\mathbf{u}=\{0.5145, 0.0539, \dots, 0.0539\}$ is much flatter and hence more $\boldsymbol{w}_j$ for $j \neq i$ will be updated in every epoch.

\subsubsection{Alleviate the Infrequent Updating Problem}

We only consider the gradient of $\mathrm{KL}(\mathbf{p},\mathbf{u})$ with respect to $\boldsymbol{w}_j$ ($j \neq i$). Note that $\mathbf{u}$ is not involved in gradient computation (in knowledge distillation~\cite{kd:hinton} the teacher predictions are not involved in gradient computation, either). Now we can get (see appendix for derivations)
\begin{equation}
\setlength\abovedisplayskip{4.5pt}
\setlength\belowdisplayskip{4.5pt}
\label{eq:Sqrt_Grad}
	 \frac{\partial \mathrm{KL}(\mathbf{p},\mathbf{u})}{\partial p_k}  = 0.5 \log p_k + (1+\log c) \,,
\end{equation}
where we define $c=\sum_s \sqrt{p_s}$. Using the above example where $p_j=0.01$, from \eqref{eq:grad} we can get:
\begin{equation}
   \left\|\frac{\partial L_{\text{CE}}} {\partial \boldsymbol{w}_j} \right\|_2 = {p_j} \left\|\boldsymbol{z}_i \right\|_2 = 0.01 \left\|\boldsymbol{z}_i\right\|_2 \,.
\end{equation}

For $L_{\text{SqrtKL}}$, we can also calculate the gradient w.r.t. $\boldsymbol{w}_j$ (see appendix for detailed derivations):

\begin{equation}
   \left\| \frac{\partial L_{\text{SqrtKL}}} {\partial \boldsymbol{w}_j} \right\|_2 \approx 0.021 \left\|\boldsymbol{z}_i\right\|_2 \,,
\end{equation}
where the update range of $\boldsymbol{w}_j$ is doubled, hence mitigating the infrequent updating problem and it will be further alleviated by increasing the coefficient $\lambda$ introduced later.

Note that NPID~\cite{memorybank:wu:CVPR18} uses proximal optimization to accelerate convergence of instance discrimination:
\begin{equation}
\setlength\abovedisplayskip{4.5pt}
\setlength\belowdisplayskip{4.5pt}
    L_p = \Vert \boldsymbol{z}_i-\boldsymbol{w}_i \Vert_2^2 \,.
\end{equation}

However, we can find that $\frac{\partial L_p}{\partial \boldsymbol{w}_j}=0$ for $j\neq i$, which means this loss does not solve the infrequent updating issue. The difference between $L_{\text{SqrtKL}}$ and $L_p$ in gradient calculation explains why our method is significantly better than NPID in the following experimental results.

\subsubsection{From an Optimization Perspective}\label{sec:optimize}

$L_{\text{SqrtKL}}$ can be decomposed into two components:
\begin{equation}
    L_{\text{SqrtKL}} = \underbrace{\sum_k p_k \log p_k}_{L_1}\underbrace{-\sum_k p_k \log u_k}_{L_2}
\end{equation}

To minimize $L_1$ amounts to maximize $-\sum_k p_k \log p_k$, or max entropy~\cite{maxentropy}. $L_1$ achieves its minimum when $p_k=\frac{1}{N}$ for all $k$. Obviously, $L_2$ achieves its minimum when
$$
p_j = \left\{
	\begin{aligned}
	1 \quad & j=\underset{k}{\arg\max u_k}  \\
	0 \quad & \text{otherwise} \\
	\end{aligned}
	\right. \,.
$$
Note that $L_2$ is determined by the largest value in the distribution, hence minimizing the cross entropy loss will in effect minimize $L_2$, too.

While $L_2$ makes the distribution sharper, $L_1$ makes it flatter. In the appendix, we show that combining $L_1$ and $L_2$ gives the best results, and $L_1$ is more important in $L_{\text{SqrtKL}}$. 

The overall loss function of our method is:
\begin{equation}
    L = L_{\text{CE}} + \lambda L_{\text{SqrtKL}} \,,
\end{equation}
where $\lambda$ is a hyper-parameter.

\begin{table*}
    \caption{Linear evaluation results on three benchmark datasets. All pretrained for 400 epochs and we report the total pretraining cost (in hours) using 4 Tesla K80 cards on CIFAR-10 as an example.}
	\label{tab:cifar-result}
	\renewcommand{\arraystretch}{0.9}
	\centering
    \small
	\setlength{\tabcolsep}{3pt}
	\begin{tabular}{c|c| c|c|c| c| c|c|c}
            \hline
            \multirow{2}{*}{Backbone}&\multirow{2}{*}{Method} &Single  & Single & Training & GPU & \multicolumn{3}{c}{Accuracy (\%)} \\
            \cline{7-9}
             &  & Crop & Network & Cost (h) & Memory (MB) & CIFAR-10 & CIFAR-100 & Tiny-ImageNet \\
            \hline
            \multirow{8}{*}{ResNet-18} & BYOL~\cite{byol:grill:NIPS20} &  $\times$ & $\times$ &11.94 & 2897 &89.3 &62.6 & 32.6 \\
            & JigClu~\cite{jigclu:CVPR21} & \checkmark& \checkmark & 11.59 & 2344& 88.7&55.3 & 33.4 \\
            &SimSiam~\cite{simsiam:kaiming:cvpr2021} & $\times$ & \checkmark &\pt7.16 &2501 &90.7 &65.5 & 37.1 \\
            &SimCLR~\cite{simclr:hinton:ICML20} & $\times$ & \checkmark & \pt6.63  & 2185 & 89.4 & 59.2 & 37.6  \\
            & MoCov2~\cite{mocov2:xinlei:arxiv2020} & $\times$ & $\times$  & \pt6.54&1757& 88.9 & 62.5 &  35.8 \\

            &PID~\cite{idmm:ECCV2022} & \checkmark & \checkmark & \pt6.53 & 3639 & 89.8& 63.6 & 36.8 \\
            & NPID~\cite{memorybank:wu:CVPR18} & \checkmark & \checkmark & \pt4.15 & 1879 & 80.8 & 50.9 & 27.3 \\
            
              &  \cellcolor{LightCyan}\textbf{Ours} & \cellcolor{LightCyan}\checkmark & \cellcolor{LightCyan}\checkmark & \cellcolor{LightCyan}\pt\textbf{3.36} & \cellcolor{LightCyan}\textbf{1715} & \cellcolor{LightCyan}\textbf{91.1} & \cellcolor{LightCyan}\textbf{67.9} & \cellcolor{LightCyan}\textbf{39.7} \\
            \hline 
            \multirow{5}{*}{MobileNetv2} & BYOL~\cite{byol:grill:NIPS20} &  $\times$ & $\times$ & 12.61  &4503  &88.1 & 61.2 & 28.7  \\
            & SimSiam~\cite{simsiam:kaiming:cvpr2021} & $\times$ & \checkmark & \pt9.36 & 4275  & 86.1 & 50.0 & 20.5 \\
            &  SimCLR~\cite{simclr:hinton:ICML20} & $\times$ & \checkmark & \pt8.95 & 4061 & \textbf{88.9} & 62.4 & 23.6 \\
            & MoCov2~\cite{mocov2:xinlei:arxiv2020} & $\times$ & $\times$ & \pt8.12 & 2599  & 83.3 & 51.6 & 21.3 \\
             &  \cellcolor{LightCyan}\textbf{Ours} & \cellcolor{LightCyan}\checkmark & \cellcolor{LightCyan}\checkmark  & \cellcolor{LightCyan}\pt\textbf{3.95}  & \cellcolor{LightCyan}\textbf{2181}  & \cellcolor{LightCyan}88.7 & \cellcolor{LightCyan}\textbf{65.5} & \cellcolor{LightCyan}\textbf{36.2} \\
            \hline 
            \multirow{5}{*}{ResNet-50}& BYOL~\cite{byol:grill:NIPS20} &  $\times$ & $\times$ &31.08 &9435 & 90.3 & 66.7 & 41.1 \\
            & SimSiam~\cite{simsiam:kaiming:cvpr2021} & $\times$ & \checkmark &22.32 &9139 & 90.9 & 64.3 & 39.3 \\
            &SimCLR~\cite{simclr:hinton:ICML20} & $\times$ & \checkmark & 21.94&8951 &91.5 & 66.2 & 42.8  \\
            & MoCov2~\cite{mocov2:xinlei:arxiv2020} & $\times$ & $\times$ & 14.72&5373 & 90.2 & 66.5 & 42.2 \\

             &  \cellcolor{LightCyan}\textbf{Ours} & \cellcolor{LightCyan}\checkmark & \cellcolor{LightCyan}\checkmark &\cellcolor{LightCyan}\textbf{10.75} & \cellcolor{LightCyan}\textbf{5095} & \cellcolor{LightCyan}\textbf{92.0} & \cellcolor{LightCyan}\textbf{71.6} & \cellcolor{LightCyan}\textbf{44.9} \\
            \hline
	\end{tabular}
\end{table*}

\section{Experimental Results}
We introduce the implementation details in Sec.~\ref{sec:details}. We experiment on CIFAR-10~\cite{cifar}, CIFAR-100~\cite{cifar}, and Tiny-ImageNet in Sec.~\ref{sec:exp-cifar}. We experiment on ImageNet~\cite{ILSVRC2012:russakovsky:IJCV15} and study the transfer performance of ImageNet pretrained models on downstream recognition, object detection, and instance segmentation benchmarks in Sec.~\ref{sec:exp-imagenet}. Finally, we investigate the effects of different components and hyper-parameters in our method in Sec.~\ref{sec:ablation}. All our experiments were conducted using PyTorch with Tesla K80 and 3090 GPUs. Codes will be publicly available upon acceptance.

\subsection{Implementation Details}\label{sec:details}
\noindent\textbf{Datasets.} The main experiments are conducted on four benchmark datasets, i.e., CIFAR-10, CIFAR-100, Tiny-ImageNet and ImageNet. Tiny-ImageNet contains 100,000 training and 10,000 validation images from 200 classes at $64\times64$ resolution. We also conduct transfer experiments on 2 recognition benchmarks as well as 2 detection benchmarks Pascal VOC 07\&12~\cite{VOC:mark:IJCV10} and COCO2017~\cite{coco:LinTY:ECCV14}.

\noindent\textbf{Backbones.} In addition to the commonly used ResNet-50~\cite{resnet:he:CVPR16} in recent SSL papers, we also adopt 4 smaller networks to study model efficiency, i.e., ResNet-18~\cite{resnet:he:CVPR16}, MobileNetv2~\cite{mobilenetv2:sabdker:CVPR18}, MobileNetv3~\cite{mobilenetv3:ICCV19}, and EfficientNet~\cite{efficientnet:icml19} for our experiments. Sometimes we abbreviate ResNet-18/50 to R-18/50, and MobileNetv3 to Mobv3.

\noindent\textbf{Training details.} We use SGD for pretraining, with a batch size of 512 and a base lr=0.1. The learning rate has a cosine decay schedule. The weight decay is 0.0001 and the SGD momentum is 0.9. We set $m=0.5$ and $\lambda=20$ and we pretrain for 400 epochs on CIFAR-10, CIFAR-100, and Tiny-ImageNet, and 200 epochs on ImageNet by default. 

\subsection{Experiments on CIFAR and Tiny ImageNet}\label{sec:exp-cifar}

We first compare our method with 4 popular dual-branch SSL methods (BYOL~\cite{byol:grill:NIPS20}, SimSiam~\cite{simsiam:kaiming:cvpr2021}, SimCLR~\cite{simclr:hinton:ICML20}, MoCov2~\cite{mocov2:xinlei:arxiv2020}) and 3 single-branch methods (PID~\cite{exemplar:alexey:nips14}, NPID~\cite{memorybank:wu:CVPR18}, Jigclu~\cite{jigclu:CVPR21}) on CIFAR-10, CIFAR-100 and Tiny-ImageNet using three CNN backbones in Table~\ref{tab:cifar-result}. All methods are pretrained for 400 epochs for fair comparisons and we report the total training hours on CIFAR-10 using 4 K80 GPUs. We also report the GPU memory usage of each method during training and here we use the same batch size 512 for fair comparisons. We report the linear probing accuracy on each dataset, following the practice in \cite{ssql:cao:ECCV2022}.

\noindent\textbf{Comparison with Dual-Branch Methods.} As shown in Table~\ref{tab:cifar-result}, our method only requires a single network branch and a single crop, thus achieving much lower memory usage and training time than mainstream dual-branch SSL methods. When compared with SimSiam~\cite{simsiam:kaiming:cvpr2021}, our method only needs 46.9\% of the training time and 68.6\% of the GPU memory usage, but achieves 0.4\%, 2.4\% and 2.6\% higher accuracy on CIFAR-10, CIFAR-100 and Tiny-ImageNet under R-18, respectively. When compared with BYOL~\cite{byol:grill:NIPS20}, our method achieves significantly higher accuracy, using only one-third of the training time and nearly half of the GPU memory usage. We can reach similar conclusions by comparing with other methods and backbones.

Note that current self-supervised methods such as MoCov2~\cite{mocov2:xinlei:arxiv2020} and SimSiam perform poorly on small architectures such as MobileNetv2, as mentioned in \cite{seed:fang:ICLR21}. In contrast, our method can also achieve very good results together with small models, especially on CIFAR-100 and Tiny-ImageNet. We think the reason for this is that the capacity of the small model is not enough to learn difficult self-supervised tasks. In contrast, our single-branch classification method is simple to learn and our proposed method makes the model easier to converge.

\begin{table*}[t]
	\caption{Downstream object detection performance on VOC 07\&12 and linear evaluation accuracy on Tiny-ImageNet when pretrained on ImageNet subsets using R-18 and R-50. Improvements compared to MoCov2 are listed in parentheses.}
	\label{tab:imagenet-subset-detection}
	\centering
	\small
	\renewcommand{\arraystretch}{0.95}
	\renewcommand{\multirowsetup}{\centering}
	\begin{tabular}{c|c|r|r|r|ccc|c}
		\hline
		 \multirow{2}{*}{Backbone} &\multicolumn{4}{c|}{Pretraining}          &           \multicolumn{3}{c|}{VOC 07\&12} &\multirow{2}{*}{Tiny-ImageNet}         \\ 
		\cline{2-8}
		 & Method & \#Images&Epochs & Cost (h)& $\text{AP}_{50}$ & $\text{AP}$ & $\text{AP}_{75}$ \\
		\hline
\multirow{9}{*}{ResNet-18}&random init.                           &0&   0&  0 &     59.2 \pt\pt\pt\pt\pt       &       32.5 \pt\pt\pt\pt\pt        &         31.5 \pt\pt\pt\pt\pt  & \pt 0.5 \pt\pt\pt\pt\pt       \\
		\cline{2-9}
		& MoCov2~\cite{mocov2:xinlei:arxiv2020}	&\multirow{2}{*}{10,000} &   \multirow{2}{*}{200}  & 0.43 &   61.8 \pt\pt\pt\pt\pt    &   34.3 \pt\pt\pt\pt\pt     &   33.4 \pt\pt\pt\pt\pt   & \pt 9.7\pt\pt\pt\pt\pt   \\
		 & \textbf{Ours} & & &\cellcolor{LightCyan}\textbf{0.28}  & \cellcolor{LightCyan}\textbf{67.1} (\textcolor{grassgreen}{+5.3}) & \cellcolor{LightCyan}\textbf{38.5} (\textcolor{grassgreen}{+4.2})& \cellcolor{LightCyan}\textbf{37.8} (\textcolor{grassgreen}{+4.4})& \cellcolor{LightCyan}\textbf{19.4} (\textcolor{grassgreen}{+9.7})  \\
		\cline{2-9}
		& MoCov2~\cite{mocov2:xinlei:arxiv2020}	&\multirow{2}{*}{10,000} &   \multirow{2}{*}{800}  & 1.72 &     65.0\pt\pt\pt\pt\pt    &    37.2\pt\pt\pt\pt\pt     &    37.0\pt\pt\pt\pt\pt  & 13.7\pt\pt\pt\pt\pt   \\
		& \textbf{Ours} & & &\cellcolor{LightCyan}\textbf{1.12}  & \cellcolor{LightCyan}\textbf{68.5} (\textcolor{grassgreen}{+3.5}) & \cellcolor{LightCyan}\textbf{39.8} (\textcolor{grassgreen}{+2.6})& \cellcolor{LightCyan}\textbf{39.8} (\textcolor{grassgreen}{+2.8})& \cellcolor{LightCyan}\textbf{20.5} (\textcolor{grassgreen}{+6.8})  \\
		\cline{2-9}
		& MoCov2~\cite{mocov2:xinlei:arxiv2020} &\multirow{3}{*}{100,000} &   \multirow{3}{*}{200}  & 4.33  & 70.6       \pt\pt\pt\pt\pt     &     41.6 \pt\pt\pt\pt\pt   &      42.7 \pt\pt\pt\pt\pt      &   23.6 \pt\pt\pt\pt\pt    \\
        &SimSiam~\cite{simsiam:kaiming:cvpr2021} &  & & 4.47 & 71.1 \pt\pt\pt\pt\pt & 42.5 \pt\pt\pt\pt\pt & 44.3 \pt\pt\pt\pt\pt & 24.3 \pt\pt\pt\pt\pt \\
		& \textbf{Ours} & &  &\cellcolor{LightCyan}\textbf{2.81}& \cellcolor{LightCyan}\textbf{71.8}  (\textcolor{grassgreen}{+1.2}) & \cellcolor{LightCyan}\textbf{43.1} (\textcolor{grassgreen}{+1.5}) & \cellcolor{LightCyan}\textbf{44.7} (\textcolor{grassgreen}{+2.0}) &  \cellcolor{LightCyan}\textbf{29.5} (\textcolor{grassgreen}{+5.9})  \\
		\cline{2-9}
		& MoCov2~\cite{mocov2:xinlei:arxiv2020}&\multirow{2}{*}{100,000} &   \multirow{2}{*}{800}  & 17.32 &        72.7  \pt\pt\pt\pt\pt     &     43.6 \pt\pt\pt\pt\pt   &       45.3 \pt\pt\pt\pt\pt      &   27.4 \pt\pt\pt\pt\pt   \\
           & \textbf{Ours} &  & &\cellcolor{LightCyan}\textbf{11.24} & \cellcolor{LightCyan}\textbf{73.4} (\textcolor{grassgreen}{+0.7}) & \cellcolor{LightCyan}\textbf{44.8} (\textcolor{grassgreen}{+1.2}) & \cellcolor{LightCyan}\textbf{47.0} (\textcolor{grassgreen}{+1.7}) & \cellcolor{LightCyan}\textbf{32.4} (\textcolor{grassgreen}{+5.0}) \\
		\hline
		\multirow{8}{*}{ResNet-50}&random init.                           &0&   0&  0 &   63.0 \pt\pt\pt\pt\pt       &     36.7 \pt\pt\pt\pt\pt        &       36.9 \pt\pt\pt\pt\pt  & \pt0.5 \pt\pt\pt\pt\pt       \\
		\cline{2-9}
		& MoCov2~\cite{mocov2:xinlei:arxiv2020}	&\multirow{2}{*}{10,000} &   \multirow{2}{*}{800}  & 1.88 &   71.6 \pt\pt\pt\pt\pt    &  43.9 \pt\pt\pt\pt\pt     &  45.9 \pt\pt\pt\pt\pt  & 23.6 \pt\pt\pt\pt\pt   \\
		& \textbf{Ours} & & &\cellcolor{LightCyan}\textbf{1.64} & \cellcolor{LightCyan}\textbf{76.8} (\textcolor{grassgreen}{+5.2})& \cellcolor{LightCyan}\textbf{49.3} (\textcolor{grassgreen}{+5.4}) & \cellcolor{LightCyan}\textbf{53.6}  (\textcolor{grassgreen}{+7.7}) & \cellcolor{LightCyan}\textbf{26.3}  (\textcolor{grassgreen}{+2.7})\\ 
		\cline{2-9}
		& MoCov2~\cite{mocov2:xinlei:arxiv2020}&\multirow{3}{*}{100,000} &   \multirow{3}{*}{200}  & 4.65  &       76.2 \pt\pt\pt\pt\pt     &     48.0 \pt\pt\pt\pt\pt   &    51.6  \pt\pt\pt\pt\pt      & 35.3   \pt\pt\pt\pt\pt    \\
              & SimSiam~\cite{simsiam:kaiming:cvpr2021} & & & 5.42 & 76.4 \pt\pt\pt\pt\pt & 49.8 \pt\pt\pt\pt\pt & 54.2 \pt\pt\pt\pt\pt & 30.5 \pt\pt\pt\pt\pt \\
		& \textbf{Ours} & & &\cellcolor{LightCyan}\textbf{4.09} & \cellcolor{LightCyan}\textbf{78.2} (\textcolor{grassgreen}{+2.0}) & \cellcolor{LightCyan}\textbf{51.1} (\textcolor{grassgreen}{+3.1}) & \cellcolor{LightCyan}\textbf{55.7} (\textcolor{grassgreen}{+4.1}) & \cellcolor{LightCyan}\textbf{36.3} (\textcolor{grassgreen}{+1.0})  \\
		\cline{2-9}
		& MoCov2~\cite{mocov2:xinlei:arxiv2020}&\multirow{2}{*}{100,000} &   \multirow{2}{*}{800}  & 18.62 &         78.7 \pt\pt\pt\pt\pt     &     51.5 \pt\pt\pt\pt\pt   &    56.3  \pt\pt\pt\pt\pt      & 43.7  \pt\pt\pt\pt\pt   \\
           & \textbf{Ours} & & &\cellcolor{LightCyan}\textbf{16.36} & \cellcolor{LightCyan}\textbf{79.7} (\textcolor{grassgreen}{+1.0}) & \cellcolor{LightCyan}\textbf{53.3} (\textcolor{grassgreen}{+1.8}) &  \cellcolor{LightCyan}\textbf{58.8} (\textcolor{grassgreen}{+2.2}) & \cellcolor{LightCyan} \textbf{44.3} 
 (\textcolor{grassgreen}{+0.6}) \\
		\hline
	\end{tabular}
\end{table*}

\begin{table*}[t]
    \caption{ImageNet (subsets) pretraining results on small architectures. All pretrained for 200 epochs and we report the linear evaluation accuracy (\%) when transferring to CIFAR-100 and the pretraining hours using 8 3090 cards. $\dagger$: Results from \cite{smallmodel:AAAI22}.}
	\label{tab:linear-eval-small-imagenet}
	\centering
 	\renewcommand{\arraystretch}{0.95}
    \small
	\setlength{\tabcolsep}{4pt}
	\begin{tabular}{c|c|c|c|c|c|c|c}
            \hline
            \multirow{2}{*}{Backbone} & \# Images & \multicolumn{2}{c|}{10,000} & \multicolumn{2}{c|}{100,000} & \multicolumn{2}{c}{1,281,167} \\
            \cline{2-8}
            &Method &Linear (\%) $\uparrow$ & Cost (h) $\downarrow$ & Linear (\%) $\uparrow$& Cost (h)  $\downarrow$&Linear (\%) $\uparrow$& Cost (h) $\downarrow$\\
            \hline
            \multirow{2}{*}{Mobv3-small (2.5M)} & MoCov2 & 21.8  & 0.42 & 33.0 &4.18 &    $40.4^\dagger$&53.55 \\
            & \cellcolor{LightCyan} \textbf{Ours} &\cellcolor{LightCyan}\textbf{34.0}  & \cellcolor{LightCyan}\textbf{0.34}& \cellcolor{LightCyan}\textbf{39.9}& \cellcolor{LightCyan}\textbf{3.43}& \cellcolor{LightCyan}\textbf{44.3}\pt& \cellcolor{LightCyan}\textbf{43.94}\\ 
            \hline
            \multirow{2}{*}{Mobv3-large (5.4M)} & MoCov2 & 28.1  & 0.42 & 32.5 & 4.23 &    $42.4^\dagger$ &54.19 \\
            & \cellcolor{LightCyan} \textbf{Ours} &\cellcolor{LightCyan}\textbf{31.5}  & \cellcolor{LightCyan}\textbf{0.38}& \cellcolor{LightCyan}\textbf{36.1}& \cellcolor{LightCyan}\textbf{3.79}& \cellcolor{LightCyan}\textbf{50.1}\pt& \cellcolor{LightCyan}\textbf{48.56}\\ 
            \hline
            \multirow{2}{*}{EfficientNet-b0 (5.3M)} & MoCov2 & 26.0  & 0.43 & 34.8 &4.31 &    $43.2^\dagger$ & 55.22\\
            & \cellcolor{LightCyan} \textbf{Ours} &\cellcolor{LightCyan}\textbf{38.1}  & \cellcolor{LightCyan}\textbf{0.39}& \cellcolor{LightCyan}\textbf{39.9}& \cellcolor{LightCyan}\textbf{3.87}& \cellcolor{LightCyan}\textbf{47.8}\pt& \cellcolor{LightCyan}\textbf{49.56} \\ 
            \hline
            \multirow{2}{*}{ResNet-18 (11.7M)} & MoCov2 & 39.9  & 0.43 & 51.7 &4.33 &    $54.0^\dagger$ & 55.47\\
            & \cellcolor{LightCyan} \textbf{Ours} 
            &\cellcolor{LightCyan}\textbf{48.8}  & \cellcolor{LightCyan}\textbf{0.28}& \cellcolor{LightCyan}\textbf{55.3}& \cellcolor{LightCyan}\textbf{2.81}& \cellcolor{LightCyan}\textbf{60.4}\pt& \cellcolor{LightCyan}\textbf{36.05}\\ 
            \hline
	\end{tabular}
\end{table*}

\noindent\textbf{Comparison with Single-Branch Methods.} Although both our method and PID~\cite{exemplar:alexey:nips14,idmm:ECCV2022} are single-branch ones, PID requires a parameterized classification layer, which brings additional training (gradient back-propagation) and storage overhead, and will inevitably deteriorate with more training data. In contrast, our method is non-parametric and the training time and storage are less affected by the amount of training data. At the same time, our corrected update rule and SqrtKL loss also enable us to achieve much better results on all three datasets than NPID~\cite{memorybank:wu:CVPR18} and PID, which are also based on instance discrimination. When compared with the state-of-the-art single-branch method JigClu~\cite{jigclu:CVPR21}, the training time of our method is reduced by 71\% for ResNet-18 (from 11.59 to 3.36 hours), because we do not need complex patch-level augmentations.

In short, our method greatly improves the training efficiency of the SSL method, achieves the best results with the least training overhead, and has a greater improvement in small models. It can be seen that among all comparison methods, MoCov2 is the strongest opponent in the tradeoff between accuracy and efficiency, so the main comparison method in our subsequent experiments will be MoCov2.

\subsection{ImageNet and Transferring Experiments}\label{sec:exp-imagenet}

In this subsection, we first perform unsupervised pretraining on the large-scale ImageNet training set without using labels, then investigate the downstream object detection performance on COCO2017~\cite{coco:LinTY:ECCV14} and Pascal VOC 07\&12~\cite{VOC:mark:IJCV10}. The detector is Faster R-CNN~\cite{faster-rcnn:ren:NIPS15} for Pascal VOC, and Mask R-CNN~\cite{mask-rcnn:he:ICCV17} for COCO, both with the C4 backbone~\cite{faster-rcnn:ren:NIPS15}, following \cite{mocov2:xinlei:arxiv2020,simsiam:kaiming:cvpr2021}. 

\begin{table*}
    \caption{Transfer Learning. All unsupervised methods are based on 200-epoch pretraining in ImageNet. We use Faster R-CNN for VOC and Mask R-CNN for COCO under the C4-backbone. Bold entries are the best two results following the style of~\cite{simsiam:kaiming:cvpr2021}. {$\dagger$}: Results from~\cite{simsiam:kaiming:cvpr2021}.}
	\label{tab:coco-result}
	\renewcommand{\arraystretch}{0.75}
	\centering
    \small
	\setlength{\tabcolsep}{2.5pt}
	\begin{tabular}{c|c|c c c|c c c|c c c|c c c}
            \multirow{2}{*}{Method} & Single &\multicolumn{3}{c|}{VOC 07 detection} & \multicolumn{3}{c|}{VOC 07+12 detection} & \multicolumn{3}{c|}{COCO detection} & \multicolumn{3}{c}{COCO instance seg.}\\
             & Branch & $\text{AP}_{50}$& $\text{AP}$ & $\text{AP}_{75}$ & $\text{AP}_{50}$& $ \text{AP}$ & $\text{AP}_{75}$ & $\text{AP}_{50}^{\text{bb}}$ & $\text{AP}^{\text{bb}}$ & $\text{AP}_{75}^{\text{bb}}$ & $\text{AP}_{50}^{\text{mask}}$ & $\text{AP}^{\text{mask}}$ & $\text{AP}_{75}^{\text{mask}}$ \\
            \hline
            $\text{scratch}^\dagger$ & - & 35.9 & 16.8 & 13.0 & 60.2 & 33.8 & 33.1 & 44.0 & 26.4 & 27.8 & 46.9 & 29.3 & 30.8 \\
            $\text{ImageNet supervised}^\dagger$ & \checkmark  & 74.4 &42.4 &42.7 & 81.3 & 53.5 & 58.8 & 58.2 & 38.2 & 41.2 & 54.7 & 33.3 & 35.2 \\
            $\text{SimCLR}^\dagger$~\cite{simclr:hinton:ICML20} & $\times$ & 75.9 & 46.8 & 50.1 & 81.8 & 55.5 & 61.4 & 57.7 & 37.9 & 40.9 & 54.6 & 33.3 & 35.3 \\
            $\text{MoCov2}^\dagger$~\cite{mocov2:xinlei:arxiv2020}&$\times$& \textbf{77.1} & \textbf{48.5} & 
            \textbf{52.5} & \textbf{82.3} & \textbf{57.0} & \textbf{63.3} & \textbf{58.8} & \textbf{39.2} & \textbf{42.5} & \textbf{55.5} & \textbf{34.3} & \textbf{36.6} \\
            $\text{BYOL}^\dagger$~\cite{byol:grill:NIPS20} &$\times$ & \textbf{77.1} & 47.0 & 49.9 & 81.4 & 55.3 & 61.1 & 57.8 & 37.9 & 40.9 & 54.3 & 33.2 & 35.0 \\
            $\text{SwAV}^\dagger$~\cite{swav:caron:NIPS20}  &$\times$& 75.5 & 46.5 & 49.6 & 81.5 & 55.4 & 61.4 & 57.6 & 37.6 & 40.3 & 54.2 & 33.1 & 35.1 \\
            $\text{SimSiam}^\dagger$~\cite{simsiam:kaiming:cvpr2021} &$\times$& 75.5 &  47.0 & 50.2 & \textbf{82.0} & 56.4 & \textbf{62.8} & 57.5 & 37.9 & 40.9 & 54.2 & 33.2 & 35.2 \\
            \rowcolor{LightCyan} \textbf{Ours} & 
            \checkmark &75.5 &\textbf{47.5}& \textbf{51.4} & \textbf{82.0} & \textbf{56.5} & 62.6 & \textbf{58.1} & \textbf{38.4} & \textbf{41.3} & \textbf{54.8} & \textbf{33.6} & \textbf{35.9} \\
            \hline
	\end{tabular}
\end{table*}

\begin{table}
    \caption{Object detection and instance segmentation results using ResNet-18 C4. $\dagger$: Results from \cite{seed:fang:ICLR21}. `T' and `$\text{AP}^{\text{mk}}$' abbreviate for pretrained teacher and `$\text{AP}^{\text{mask}}$', respectively.}
	\label{tab:coco-result-r18}
	\renewcommand{\arraystretch}{0.8}
	\setlength{\tabcolsep}{0.1pt}
	\centering
    \small
	\setlength{\tabcolsep}{2.5pt}
	\begin{tabular}{c|c|c c c|c c c}
            \multirow{2}{*}{Method}  & \multirow{2}{*}{T} & \multicolumn{3}{c|}{COCO detection} & \multicolumn{3}{c}{COCO instance seg.}\\
             & & $\text{AP}_{50}^{\text{bb}}$ & $\text{AP}^{\text{bb}}$ & $\text{AP}_{75}^{\text{bb}}$ & $\text{AP}_{50}^{\text{mk}}$ & $\text{AP}^{\text{mk}}$ & $\text{AP}_{75}^{\text{mk}}$ \\
            \hline
            $\text{MoCov2}^\dagger$~\cite{mocov2:xinlei:arxiv2020} & $\times$  & 53.9 & 35.0 & 37.7 & 51.1 & 31.0 & 33.1 \\
            $\text{SEED}^\dagger$~\cite{seed:fang:ICLR21} & R-50 & 54.2 & \textbf{35.3} & 37.8 & 51.1 & 31.1 & 33.2 \\
            $\text{SEED}^\dagger$~\cite{seed:fang:ICLR21} & R-101 & \textbf{54.3} & \textbf{35.3} & \textbf{37.9} & \textbf{51.3} & 31.3 & 33.4 \\
             \cellcolor{LightCyan} \textbf{Ours} & \cellcolor{LightCyan} $\times$ &\cellcolor{LightCyan}54.2 & \cellcolor{LightCyan}35.2 & \cellcolor{LightCyan}\textbf{37.9} & \cellcolor{LightCyan}51.2 & \cellcolor{LightCyan}\textbf{31.4} & \cellcolor{LightCyan}\textbf{33.5} \\
            \hline
	\end{tabular}
\end{table}

\noindent{\textbf{Data Efficiency.}} In order to study the data efficiency of different methods, we first compare the performance under different
data volumes by sampling the original ImageNet to smaller subsets. We randomly
sample (without using any image label) 10 thousand (10k) and 100 thousand
(100k) images to construct IN-10k and IN-100k, respectively. We only change the amount of data here and other training settings remain the same as before. 

We experiment with ResNet-18 and ResNet-50 on ImageNet subsets in Table~\ref{tab:imagenet-subset-detection} and transfer the pretrained weights to Pascal VOC 07\&12 for object detection and to Tiny-ImageNet for linear evaluation. As Table~\ref{tab:imagenet-subset-detection} shows, our method achieves significant improvements on both
downstream tasks. Take R-18 as an example, when both are trained for 200 epochs on IN-100k (100,000 images), our method is significantly better than the baseline counterpart MoCov2:
up to +1.2 $\text{AP}_{50}$, +1.5 AP, +2.0 $\text{AP}_{75}$ on VOC 07\&12 and +5.9\% accuracy on Tiny-ImageNet, with 35.1\% reduction in training time. When the amount of training data is further reduced to 10,000, our advantages will be further expanded: up to +5.3 $\text{AP}_{50}$ on VOC and +9.7\% accuracy on Tiny-ImageNet. Note that the results of our method trained for 200 epochs on IN-10k even surpass the results of MoCov2 trained for 800 epochs on IN-100k for R-50. Moreover, when comparing the results of R-18 and R-50, we find that our method will have a greater relative improvement on the smaller model R-18, especially on the linear evaluation metric of Tiny-ImageNet. These results demonstrate the training, model, and data efficiency of our method, which improves performance while reducing the training time, and has greater advantages for small data and small models (i.e., resource-constrained scenarios).

\noindent{\textbf{Model Efficiency}}. In order to further study the performance of small models, we conduct experiments with different small models on ImageNet subsets (including the entire ImageNet) in Table~\ref{tab:linear-eval-small-imagenet}. We transfer the learned representations to CIFAR-100 and conduct linear probing for comparison. Our method achieves higher accuracy than MoCov2 consistently under different lightweight backbones using training images at different scales, with less training costs. Moreover, we can see that our method's advantages are more obvious when the amount of data is reduced. Take MobileNetv3-small as an example, the improvement of our method is 3.9\% when trained on ImageNet, and it increases to 6.9\% on IN-100k and 12.2\% on IN-10k.

Then, we present ImageNet and transferring results and we use ResNet-18 and Resnet-50 as the backbone to compare with mainstream methods. We will discuss the results of linear evaluation on ImageNet later in Sec.~\ref{sec:future} and here we present the results of transferring to detection.

In Table~\ref{tab:coco-result}, we compare the learned representations of ResNet-50 on ImageNet by transferring them to other tasks, including VOC object detection and COCO object detection and instance segmentation. All methods are based on 200-epoch pretraining on ImageNet using the reproduction of SimSiam~\cite{simsiam:kaiming:cvpr2021}.
Table~\ref{tab:coco-result} shows that our method’s representations are transferable beyond the ImageNet task and it is competitive among these leading methods. SimSiam~\cite{simsiam:kaiming:cvpr2021} conjectures that the common siamese structure is a core factor for the general success of these methods while our method achieves comparable results without using a siamese network. In Table~\ref{tab:coco-result-r18}, we compare the learned representations of ResNet-18 on ImageNet by transferring them to detection and segmentation tasks. Our method achieves better results than MoCov2 and is even comparable to SEED~\cite{seed:fang:ICLR21} (which uses extra knowledge distillation).

\subsection{Ablation Study}\label{sec:ablation}

\begin{figure*}[t]
	\centering
	\subfloat[Training Loss]{
	\label{fig:trainloss}
	\includegraphics[width=0.3\linewidth]{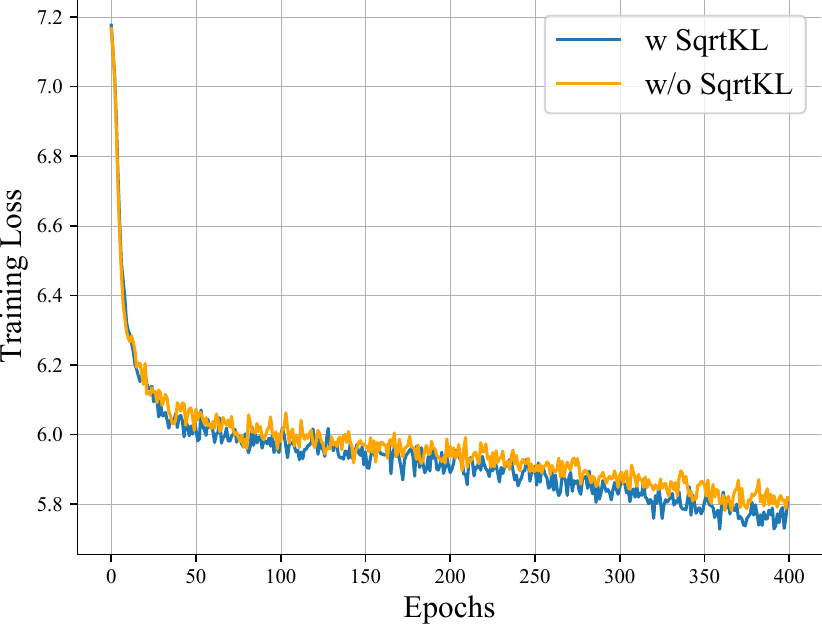}}
	\subfloat[Training Accuracy]{
	\label{fig:trainacc}
	\includegraphics[width=0.3\linewidth]{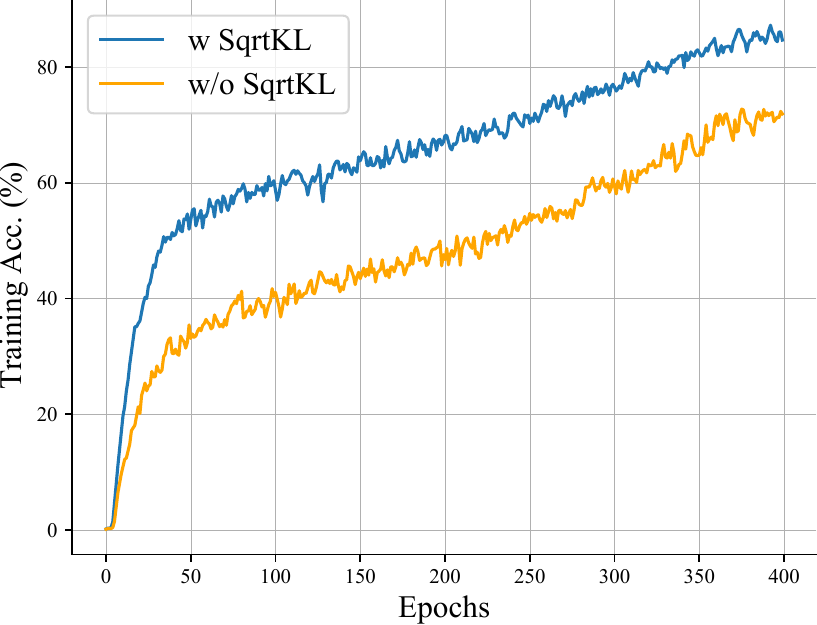}}
	\subfloat[Linear Evaluation]{
	\label{fig:linear}
	\includegraphics[width=0.3\linewidth]{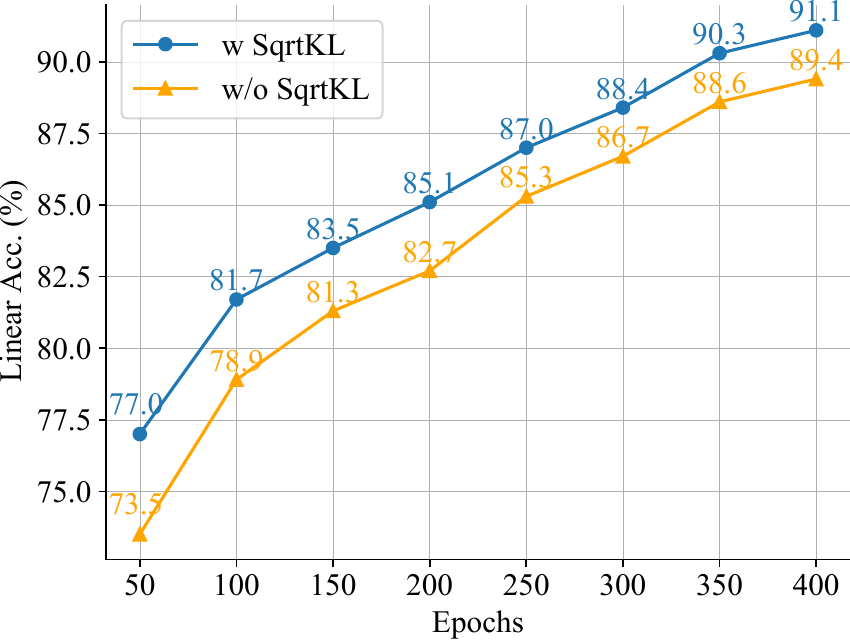}}
	\caption{Our method with vs. without SqrtKL on CIFAR-10.}
	\label{fig:SqrtKL}
\end{figure*}

\begin{figure*}[t]
    \centering
        \begin{minipage}{0.4\linewidth}
		\centering
		\small
	\setlength{\tabcolsep}{2pt}
        \begin{tabular}{c|c|c|c|c}
            Feature  & \multirow{2}{*}{SqrtKL} & Grad &  \multirow{2}{*}{CIFAR-10} &  \multirow{2}{*}{Tiny-IN} \\
             Calibrate& & Update & \\
            \hline
             $\times$ & $\times$ & $\times$ & 88.8 & 35.8 \\
             $\checkmark$ & $\times$ & $\times$ & 89.4 & 36.9 \\
             $\checkmark$ & $\times$ & $\checkmark$ & 90.0 & 37.5 \\
             $\checkmark$ & $\checkmark$ & $\times$ & \textbf{91.1} & 38.9 \\
             \rowcolor{LightCyan}$\checkmark$ & $\checkmark$ & $\checkmark$ & \textbf{91.1} & \textbf{39.7}\\
            \hline
	\end{tabular}
        \captionof{table}{Ablation study under ResNet-18.}
        \label{tab:ablation}
	\end{minipage}
        \hfill
	\begin{minipage}{0.28\linewidth}
		\centering
		\setlength{\abovecaptionskip}{0.28cm}
		\includegraphics[width=\linewidth]{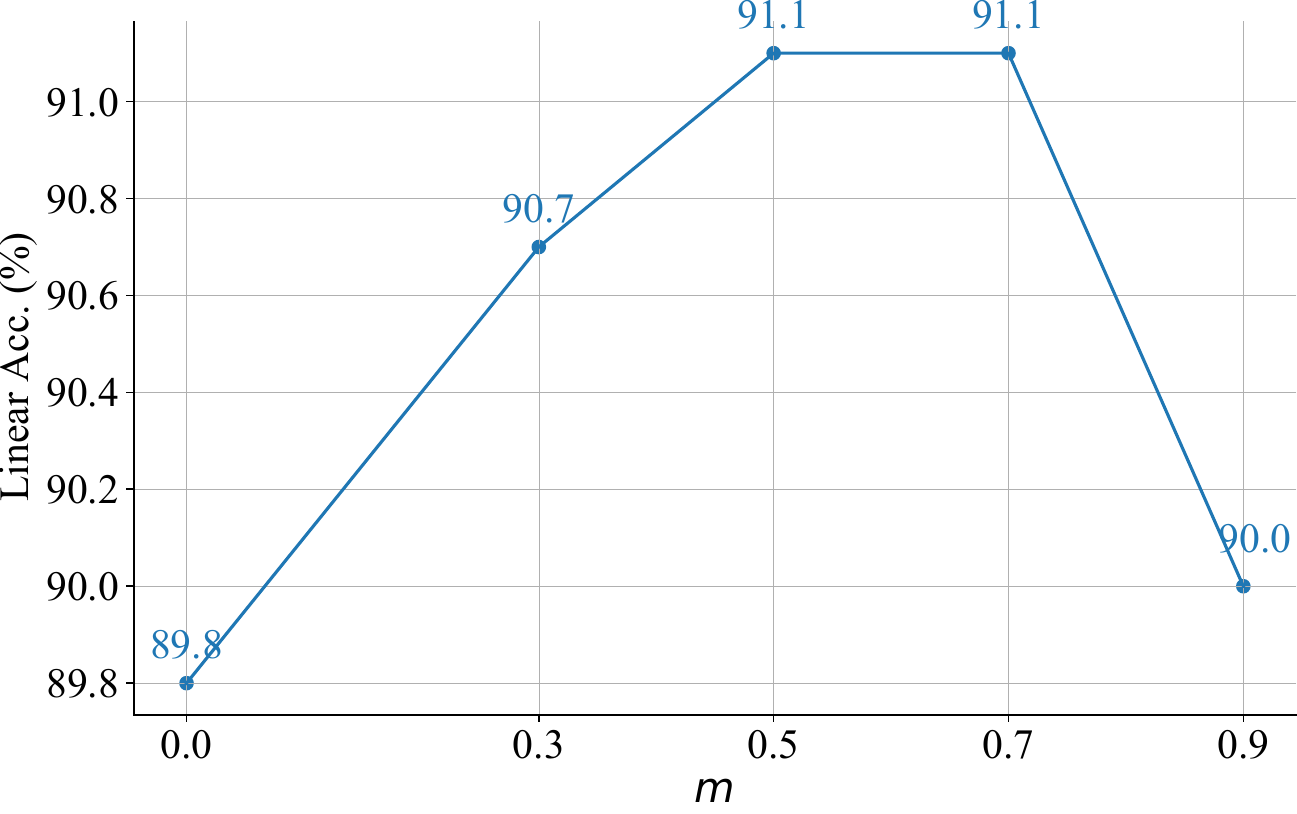}
		\caption{Effect of $m$ on CIFAR-10 under ResNet-18.}
		\label{fig:m}
	\end{minipage}
	\hfill
        \begin{minipage}{0.28\linewidth}
		\centering
		\setlength{\abovecaptionskip}{0.28cm}
		\includegraphics[width=\linewidth]{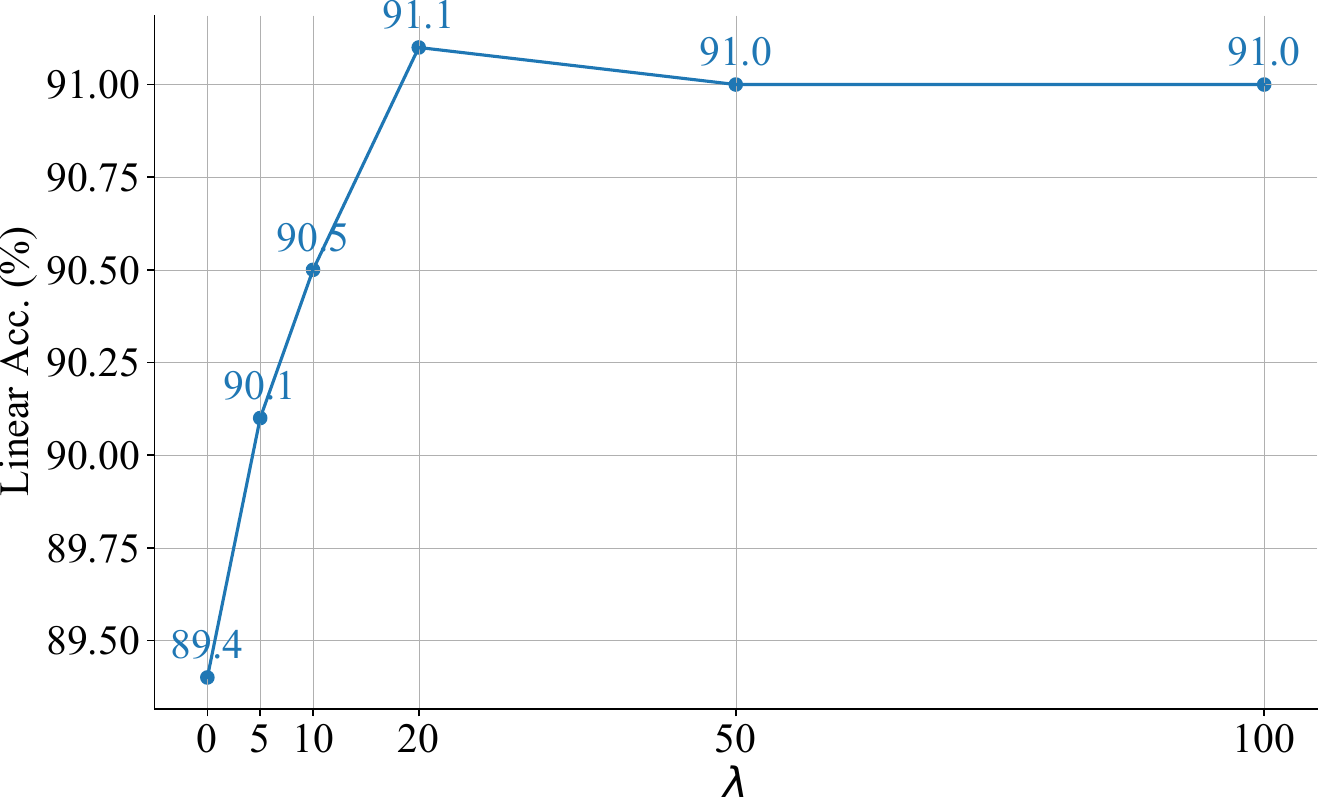}
		\caption{Effect of $\lambda$ on CIFAR-10 under ResNet-18.}
		\label{fig:lambda}
	\end{minipage}
\end{figure*}


\noindent\textbf{Effect of Feature Calibrate.} From Table~\ref{tab:ablation} we can see that this initialization brings 1.1\% gains on Tiny-ImageNet, with negligible cost (less than a minute).

\noindent\textbf{Effect of SqrtKL.} In Fig.~\ref{fig:SqrtKL} we plot the training curve of our method with and without using SqrtKL. As seen from Fig.~\ref{fig:trainacc}, SqrtKL can greatly speed up the convergence of the instance classification task and our method achieves much higher instance discrimination accuracy. Moreover, from the linear accuracy comparison of each epoch in Fig.~\ref{fig:linear} (also Table~\ref{tab:ablation}), we can see that our SqrtKL can also improve the representation ability of self-supervised models. 

\noindent\textbf{Effect of Grad Update.} From Table~\ref{tab:ablation} we can see that our corrected rule \eqref{eq:momentum-update} brings 0.6\% and 0.8\% accuracy gains on Tiny-ImageNet without and with SqrtKL, respectively. Note that all our three strategies are beneficial and combining the three strategies achieves the best performance.

\noindent\textbf{Effect of Hyper-parameter $m$.} Now we study the effect of the hyper-parameter $m$, i.e., the momentum coefficient of \eqref{eq:momentum-update}. We train on CIFAR-10 for 400 epochs for all settings and the results are shown in Figure~\ref{fig:m}. We can observe that $m=0.5$ and $m=0.7$ achieve the highest accuracy. Notice that when $m=0.0$, \eqref{eq:momentum-update} is equivalent to directly updating with the results of the current iteration, that is, forgetting the previous results, so the effect is not good. It is worth mentioning that $m$ in MoCov2 and BYOL is usually set to 0.99 or 0.999, which is larger than $m=0.5$ in our paper. It is because they act on the model weights while we only act on the output features, and the update frequency of the model weights is much more frequent than features (the model weights will update in multiple iterations per epoch while each instance's representation only updates once).

\noindent\textbf{Effect of Hyper-parameter $\lambda$.} Now we study the effect of the hyper-parameter $\lambda$, i.e., the coefficient of $L_{\text{SqrtKL}}$. We train on CIFAR-10 for 400 epochs and the results are shown in Figure~\ref{fig:lambda}. We can observe that as $\lambda$ grows, the accuracy steadily improves and will not continue to improve when it grows beyond 20. Notice that we directly set $\lambda$ to 20 for all
our experiments throughout this paper and did not tune it under different datasets or backbones. It also indicates that we can get better results with more carefully tuned $\lambda$. 

\begin{table}
    \caption{ImageNet linear evaluation accuracy (\%) of different methods under ResNet-50.}
	\label{tab:linear-eval}
	\renewcommand{\arraystretch}{0.75}
	\centering
    \small
	\setlength{\tabcolsep}{3.5pt}
	\begin{tabular}{c|c|c}
            \hline
            Method & Single Branch & Accuracy (\%) \\
            \hline
            Colorization~\cite{colorization:richard:ECCV16} & \multirow{10}{*}{\checkmark} & 39.6 \\
            JigPuz~\cite{jiasaw:mehdi:ECCV16} & &  45.7\\
            DeepCluster~\cite{deepclustering:caron:ECCV18} && 48.4\\
            NPID~\cite{memorybank:wu:CVPR18} && 54.0\\
            BigBiGan~\cite{bigbiggan:NIPS19} && 56.6\\
            LA~\cite{LocalAgg:CVPR19} && 58.8\\
            SeLa~\cite{sela:iclr20} && 61.5 \\
            CPCv2~\cite{cpcv2:icml20} && 63.8\\
            JigClu~\cite{jigclu:CVPR21} &&  66.4 \\
            \rowcolor{LightCyan} Ours &  & 64.5 \\ 
            \hline
            MoCo~\cite{moco:kaiming:CVPR20} &\multirow{5}{*}{$\times$}& 60.6 \\
            PIRL~\cite{PIRL:CVPR20} && 63.6  \\
            SimCLR~\cite{simclr:hinton:ICML20} && 64.3  \\
            PCL~\cite{PCL:iclr21} && 65.9  \\
            MoCov2~\cite{mocov2:xinlei:arxiv2020} && 67.7 \\
            \hline
	\end{tabular}
\end{table}

\section{Conclusions}\label{sec:future}

In this paper, we proposed to improve the efficiency of self-supervised learning from three aspects: algorithm, model, and data. As a solution, we proposed an efficient single-branch method based on non-parametric instance discrimination, with enhanced update rule and self-distillation loss. Various experiments show that
our method obtained a significant edge over baseline counterparts with much less training cost. Moreover, we achieved impressive results with limited amounts of training data and lightweight models, which demonstrates the model and data efficiency of our method. In the future, we will try to optimize the performance of our method on larger-scale datasets, which is a limitation of the current method. 

Despite performing well on detection, our metrics on ImageNet linear evaluation are not as good as the current mainstream dual-branch methods for ResNet-50, as shown in Table~\ref{tab:linear-eval}. This is partly because linear evaluation sometimes does not accurately measure the performance of SSL methods, as noted in~\cite{MAE:CVPR22}. More importantly, we conjecture the capacity of our method is not enough to model larger-scale data, such as ImageNet-21k. Therefore, in this paper, we mainly focused on the efficiency improvement, especially on small model and small data. Making our method suit large-scale data is an interesting future work.

\clearpage
{\small
\bibliographystyle{ieee_fullname}
\bibliography{egbib}
}

\clearpage
\appendix

\section{More Discussions about $L_{\text{SqrtKL}}$}

\subsection{Detailed Derivations}
First, we present the derivation of \eqref{eq:Sqrt_Grad} as below:

\begin{align}
	\frac{\partial \mathrm{KL}(\mathbf{p},\mathbf{u})}{\partial p_k} &= \frac{\partial \sum_s p_s\log\frac{p_s}{u_s}}{\partial p_k} \\
	 &= \frac{\partial \sum_s p_s \log p_s}{\partial p_k} - \frac{\partial \sum_s p_s \log u_s}{\partial p_k} \label{eq:1}\\
	 &= 1 + \log p_k -\log u_k \label{eq:2}\\
	 &= 1 + \log p_k - \log \sqrt{p_k} + \log \left( \sum_s \sqrt{p_s} \right) \\
	 &= 0.5 \log p_k + (1+\log c) \,,
\end{align}
where we define $c=\sum_s \sqrt{p_s}$ and use the fact that $\mathbf{u}$ is not involved in gradient computation from \eqref{eq:1} to \eqref{eq:2}. It is obvious that $c \ge 1$ because $c^2 \ge \sum_k p_k = 1$. Equally obvious is that $c \le \sqrt{N}$ --- hence $1 \le c \le \sqrt{N}$.

Then, we denote $O_k=\frac{\partial \mathrm{KL}(\mathbf{p},\mathbf{u})}{\partial p_k}$ for simplicity and calculate the gradient of $L_{\text{SqrtKL}}$ with respect to $\boldsymbol{w}_j$ ($j \neq i$):
\begin{align}
    \frac{\partial \mathrm{KL}(\mathbf{p},\mathbf{u})}{\partial \boldsymbol{w}_j} &= \sum_k \frac{\partial \mathrm{KL}(\mathbf{p},\mathbf{u})}{\partial p_k} \cdot \frac{\partial p_k}{\partial \boldsymbol{w}_j} \\
    &= \sum_{k\neq j} O_k \cdot \frac{\partial p_k}{\partial \boldsymbol{w}_j} + O_j \cdot \frac{\partial p_j}{\partial \boldsymbol{w}_j} \label{eq:17} \\
    & = \left(-\sum_{k\neq j} O_k p_k p_j + O_j (p_j-p_j^2)\right) \boldsymbol{z}_i \label{eq:18} \,,
\end{align}
where we use the equation below from \eqref{eq:17} to \eqref{eq:18}
\begin{equation}
    \frac{\partial p_k}{\partial \boldsymbol{w}_j} = p_k (\delta_{\{k=j\}}-p_j) \boldsymbol{z}_i \,.
\end{equation}
    
Then we continue to use the example in the paper, i.e., $N=10$ and $\mathbf{p}=\{0.91,0.01,\dots,0.01\}$. For $L_{\text{CE}}$, from \eqref{eq:grad} we can get:

\begin{equation}
   \frac{\partial L_{\text{CE}}} {\partial \boldsymbol{w}_j} = {p_j}\boldsymbol{z}_i = 0.01 \boldsymbol{z}_i \,.
\end{equation}

For $L_{\text{SqrtKL}}$, we can also calculate the gradient w.r.t. $\boldsymbol{w}_j$ from \eqref{eq:18} after numerical substitution:

\begin{equation}
   \frac{\partial L_{\text{SqrtKL}}} {\partial \boldsymbol{w}_j} = \approx -0.021 \boldsymbol{z}_i \,,
\end{equation}
where the update range of $\boldsymbol{w}_j$ has been expanded by over two times. Hence, we can see how $L_{\text{SqrtKL}}$ alleviate the infrequent updating problem by giving more gradients to $\boldsymbol{w}_j$ ($j\neq i$) and it will be further alleviated as we increase the coefficient $\lambda$.

\subsection{Ablation Study on $L_{\text{SqrtKL}}$}
In Sec.~\ref{sec:optimize} we analyzed that our proposed $L_{\text{SqrtKL}}$ can be decomposed into two components:
\begin{equation}
    L_{\text{SqrtKL}} = \underbrace{\sum_k p_k \log p_k}_{L_1}\underbrace{-\sum_k p_k \log u_k}_{L_2} \,,
\end{equation}
where $L_2$ makes the distribution sharper while $L_1$ makes the distribution flatter. To further demonstrate the effectiveness of our method, we experiment with only $L_1$ or $L_2$, noting that all these variants use $L_{\text{CE}}$. As shown in Table~\ref{tab:ablation-sqrtkl}, we can find that only using $L_1$ (i.e., maximizing entropy) can achieve good results. Note that $L_1$ can also alleviate the infrequent updating problem and make the distribution flatter. We can see that combining $L_1$ and $L_2$ can get better results, and $L_1$ plays a more important role in $L_{\text{SqrtKL}}$.

\begin{table}
    \caption{Ablation study on $L_{\text{SqrtKL}}$.}
    \label{tab:ablation-sqrtkl}
    \centering
    \begin{tabular}{c|c}
            Loss Formulation &  CIFAR-10 \\
            \hline
              - & 88.8 \\
              $L_1=\sum_k {p_k \log p_k}$ & 90.7  \\
              $L_2=-\sum_k {p_k \log u_k}$ & 89.8 \\
              \rowcolor{LightCyan} $L_{\text{SqrtKL}}=L_1+L_2$& \textbf{91.1} \\
            \hline
\end{tabular}

\end{table}

\begin{figure*}[t]
    \centering
    \includegraphics[width=\linewidth]{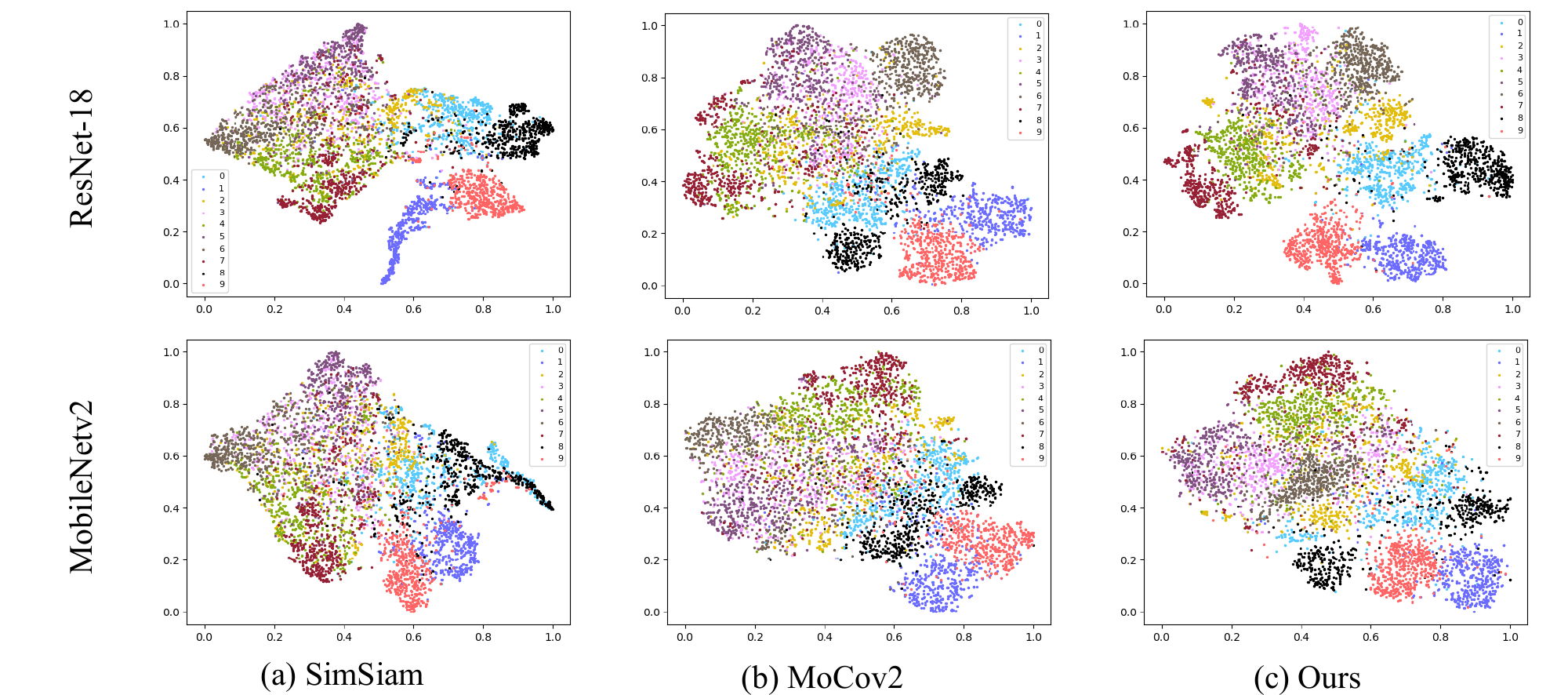}
    \caption{t-SNE~\cite{tsNE} visualization of CIFAR-10 using ResNet-18. The column (a), (b) and (c) show the results of SimSiam, MoCov2 and our method, respectively. This figure is best viewed in color.}
    \label{fig:tnse}
\end{figure*}

\section{t-SNE Visualization}
To demonstrate the effectiveness of the proposed method in a more intuitive way, we visualize the feature spaces learned by different methods in Fig.~\ref{fig:tnse}. First,
three models are trained on the CIFAR-10 dataset by using SimCLR, SimSiam
and our method, respectively. After that, 5,000 samples in CIFAR-10 are represented
accordingly and then are reduced to a two-dimensional space by t-SNE~\cite{tsNE}. As
seen, the samples are more separable in the feature space learned by our method than
both MoCov2 and SimSiam (especially under MobileNetv2).

\section{ImageNet Subsets Experiments}

\begin{figure*}[t]
    \centering
     \includegraphics[width=\linewidth]{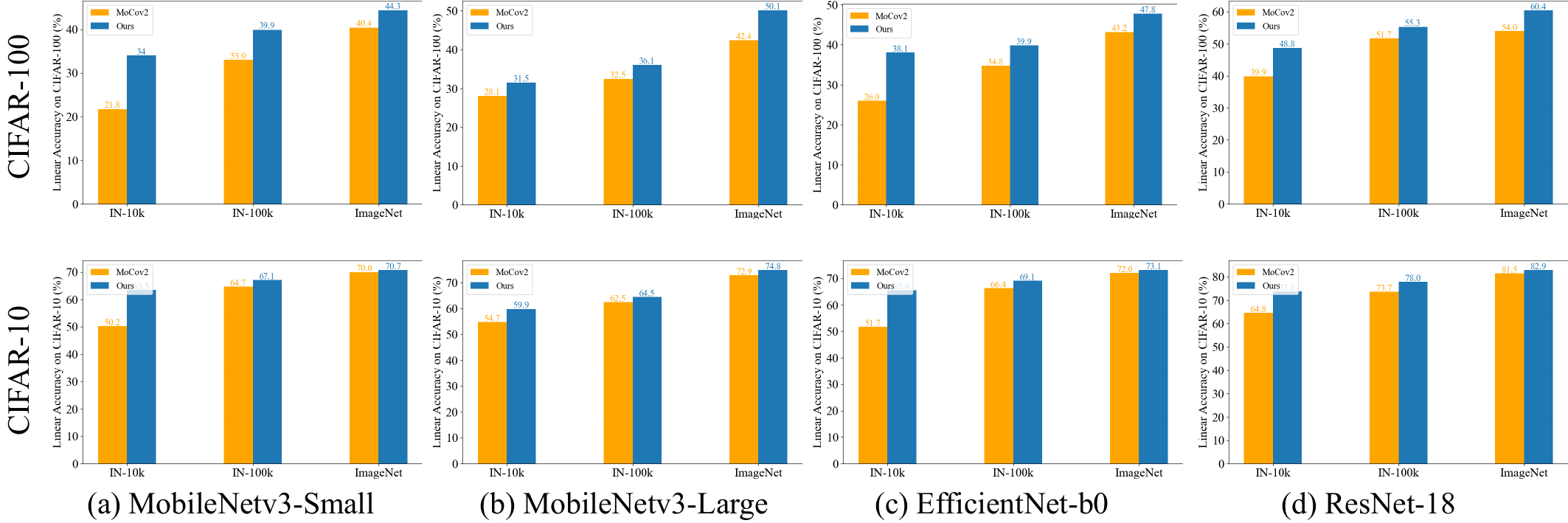}
     \vspace{-10pt}
    \caption{Comparison of our method and MoCov2 when pretrained on ImageNet subsets and then transferred to downstream recognition datasets. Upper row: Transferring to CIFAR-100. Bottom row: Transferring to CIFAR-10.}
    \label{fig:app-ImageNetsubset}
\end{figure*}

As a supplement to Table~\ref{tab:linear-eval-small-imagenet} in Sec.~\ref{sec:exp-imagenet}, we transfer the learned representations on ImageNet subsets to CIFAR-10 and we report the linear probing accuracy on CIFAR-10 for comparison in Table~\ref{tab:linear-eval-cifar10-small-imagenet}. For better illustration, we also visualize these results in Fig.~\ref{fig:app-ImageNetsubset}. We can reach similar conclusions as in the paper: 
\squishlist
\item Our method outperforms baseline counterpart MoCov2 consistently using different backbones and different scales of training images, with less training cost.
\item Our method's advantages are more obvious when the amount of data is reduced. Take MobileNetv3-small as an example, the improvement of our method is 0.7\% when trained on ImageNet, and it increases to 2.4\% on IN-100k and 13.3\% on IN-10k.
\item The amount of data required is positively correlated with the capacity of the model. Take Mobv3-small and Mobv3-large as an example, we can see that Mobv3-small even achieves better performance than Mobv3-large on IN-10k and IN-100k. It indicates that when the capacity of the model is small (i.e., has fewer parameters), a small amount of training data is enough, and the benefits brought by increasing the amount of data will become smaller and smaller. On the contrary, when the capacity of the model is large, the benefit of increasing the amount of data will be greater than that of the small model.

\squishend

\begin{table*}
    \caption{ImageNet (subsets) pretraining results on small architectures. All pretrained for 200 epochs and we report the linear evaluation accuracy (\%) when transferring to CIFAR-10 and the pretraining hours using 8 3090 cards. $\dagger$: Results from \cite{smallmodel:AAAI22}.}
	\label{tab:linear-eval-cifar10-small-imagenet}
	\centering
	\begin{tabular}{c|c|c|c|c|c|c|c}
            \hline
            \multirow{2}{*}{Backbone} & \# Images & \multicolumn{2}{c|}{10,000} & \multicolumn{2}{c|}{100,000} & \multicolumn{2}{c}{1,281,167} \\
            \cline{2-8}
            &Method &Linear (\%) $\uparrow$ & Cost (h) $\downarrow$& Linear (\%) $\uparrow$& Cost (h) $\downarrow$ &Linear (\%) $\uparrow$& Cost (h) $\downarrow$\\
            \hline
            \multirow{2}{*}{Mobv3-small (2.5M)} & MoCov2 &  50.2  & 0.42 & 64.7  &4.18 &    $70.0^\dagger$&53.55 \\
            & \cellcolor{LightCyan} \textbf{Ours} &\cellcolor{LightCyan}\textbf{63.5}  & \cellcolor{LightCyan}\textbf{0.34}& \cellcolor{LightCyan}\textbf{67.1}& \cellcolor{LightCyan}\textbf{3.43}& \cellcolor{LightCyan}\textbf{70.7}\pt& \cellcolor{LightCyan}\textbf{43.94}\\ 
            \hline
            \multirow{2}{*}{Mobv3-large (5.4M)} & MoCov2 & 54.7  & 0.42 & 62.5 & 4.23 &    $72.9^\dagger$ &54.19 \\
            & \cellcolor{LightCyan} \textbf{Ours} &\cellcolor{LightCyan}\textbf{59.9}  & \cellcolor{LightCyan}\textbf{0.38}& \cellcolor{LightCyan}\textbf{64.5}& \cellcolor{LightCyan}\textbf{3.79}& \cellcolor{LightCyan}\textbf{74.8}\pt& \cellcolor{LightCyan}\textbf{48.56}\\ 
            \hline
            \multirow{2}{*}{EfficientNet-b0 (5.3M)} & MoCov2 &  51.7 & 0.43 & 66.4 &4.31 &    $72.0^\dagger$ & 55.22\\
            & \cellcolor{LightCyan} \textbf{Ours} &\cellcolor{LightCyan}\textbf{65.6}  & \cellcolor{LightCyan}\textbf{0.39}& \cellcolor{LightCyan}\textbf{69.1}& \cellcolor{LightCyan}\textbf{3.87}& \cellcolor{LightCyan}\textbf{73.1}\pt& \cellcolor{LightCyan}\textbf{49.56} \\ 
            \hline
            \multirow{2}{*}{ResNet-18 (11.7M)} & MoCov2 & 64.8   & 0.43 &  73.7 &4.33 &    $81.5^\dagger$ & 55.47\\
            & \cellcolor{LightCyan} \textbf{Ours} 
            &\cellcolor{LightCyan}\textbf{73.8}  & \cellcolor{LightCyan}\textbf{0.28}& \cellcolor{LightCyan}\textbf{78.0}& \cellcolor{LightCyan}\textbf{2.81}& \cellcolor{LightCyan}\textbf{82.9}\pt& \cellcolor{LightCyan}\textbf{36.05}\\ 
            \hline
	\end{tabular}
\end{table*}

\end{document}